\documentclass[journal]{IEEEtran}
\usepackage{cite}

\ifCLASSINFOpdf
  \usepackage[pdftex]{graphicx}
  \graphicspath{{../images/}}
   \DeclareGraphicsExtensions{.pdf,.jpeg,.png}
\else
  \usepackage[dvips]{graphicx}
  \graphicspath{{../images/}}
   \DeclareGraphicsExtensions{.eps}
\fi
\usepackage{amsmath}
\usepackage{amssymb}
\usepackage{xcolor}
\usepackage{amsthm}

\theoremstyle{definition}

\newtheorem{rem}{Remark}
\newtheorem{assu}{Assumption}
\newtheorem{prob}{Problem}

\usepackage{algorithm}
\usepackage{algorithmic}
\usepackage{float}
\usepackage{lipsum}

\begin{document}
%
\title{Human Behavior Modeling via Identification of Task Objective and Variability}
%
%
%

\author{Sooyung~Byeon,
        Dawei~Sun,
        and~Inseok~Hwang,
\thanks{S. Byeon, D. Sun, and I. Hwang are with School of Aeronautics and Astronautics, Purdue University, West Lafayette,
Indiana, 47907, USA. e-mail: {\tt \{sbyeon,sun289,ihwang\}@purdue.edu.}}
\thanks{Manuscript received October 00, 0000; revised November 00, 0000.}}

\maketitle

\begin{abstract}

Human behavior modeling is important for the design and implementation of human-automation interactive control systems. In this context, human behavior refers to a human's control input to systems. We propose a novel method for human behavior modeling that uses human demonstrations for a given task to infer the unknown task objective and the variability. The task objective represents the human's intent or desire. It can be inferred by the inverse optimal control and improve the understanding of human behavior by providing an explainable objective function behind the given human behavior. Meanwhile, the variability denotes the intrinsic uncertainty in human behavior. It can be described by a Gaussian mixture model and capture the uncertainty in human behavior which cannot be encoded by the task objective. The proposed method can improve the prediction accuracy of human behavior by leveraging both task objective and variability. The proposed method is demonstrated through human-subject experiments using an illustrative quadrotor remote control example.

\end{abstract}




%
\IEEEpeerreviewmaketitle


\section{Introduction}

\IEEEPARstart{H}{uman} behavior modeling has been widely investigated in many applications such as driver assistance systems \cite{Driver_Assist,Driver_Assist2,Driver_Intent}, human-robot collaborated tasks \cite{Policy_Blending,Robot_Collaboration,Human_Performance_Modeling}, remotely piloted systems \cite{human_Model_Remote,My_SMC}, and unmanned aircraft systems \cite{MAV_Shared_Control}. In these applications, human behavior means a human's control input to systems. The systems refer to target platforms operated by human operators such as automobiles, robots, and quadrotors. Human behavior needs to be modeled and predicted to enable automation to assist a human without conflicting with the human's intent \cite{Topology,Robot_Human_Intent,Novice_Demonstrations}. An effective human behavior model can provide high-quality information to automation by observing and analyzing demonstrated human behaviors \cite{Human_Performance_Modeling}. Thus, human behavior models are crucial for effective human-automation interaction.

Various human behavior modeling techniques have been proposed for human-automation interactive control systems. Some techniques are applied to parameterize the human behavior model according to the given systems and environment, using basis functions \cite{Human_Modeling1,Human_Modeling2}. In the field of driving assistance, the human-in-the-loop steering dynamics has been commonly constructed using model-based parameter identification approaches \cite{Driver_Assist,Data_driven_Shared_Steering,Review_Shared_Control}. The steering torque of the human is modeled as a feedback and feedforward controller of lateral position and look-ahead point with unknown modeling parameters. Parameters are determined by human-subject experiments. However, these models heavily rely on a pre-defined structure, which may not be available for a generic human behavior model.

In the field of robotics, the learn-from-demonstrations (LfD) or imitation learning methods have been widely considered to train a behavior model from the human demonstrations \cite{Imitation_learning}. Probabilistic imitation learning approaches can account for the stochastic properties and uncertainties of the human behaviors \cite{ProMPs,Human_Behavior_Uncertainty}. These methods take multiple human demonstrations as the training data to learn high-dimensional movements which are combined to model complicated behaviors. Probabilistic modeling techniques provide a trajectory-level abstraction or an action-state-level abstraction of human behavior \cite{Imitation_learning}. However, these two abstractions only reproduce a resultant trajectory or human behaviors without any explicit reasoning underneath the observed human behaviors.

A task-objective-level abstraction can interpret the demonstrated human behavior and it provides a higher-level understanding of human behavior modeling. The task-objective-level abstraction can be learned by the inverse optimal control (IOC) or the inverse reinforcement learning (IRL) approaches \cite{IOC_incomplete,IRL,Human_Objective_Learning,IOC_IRL_Review}. The IOC has been applied to human motion characterization in neuroscience and biomechanics fields, and its efficacy has been validated by human-subject experiments \cite{ILQR,IOC_Jumping,IOC_Squatting,IOC_Human_Locomotion,IOC_Human_Motor}. However, these IOC methods cannot identify the parameters related to stochastic behaviors. The maximum entropy IRL has been proposed to address the stochastic or near-optimal properties of the given human demonstrations, but it only parameterizes the task-objective-level abstraction \cite{IRL_MaxEnt}.

\begin{figure}[!tb]
\centering
\includegraphics[width=0.48\textwidth]{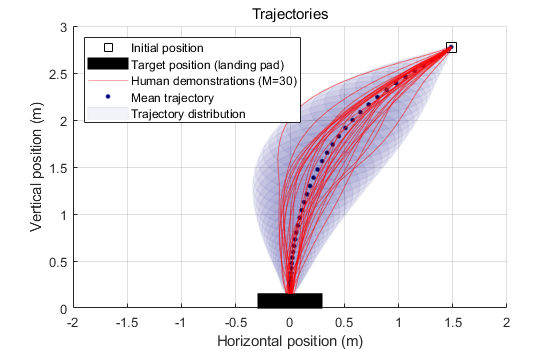}
\caption{The variability in a quadrotor landing scenario; a human operator demonstrated the landing scenario for 30 times ($M=30$) with the same initial condition, but the demonstrated trajectories are varying for each trial due to the variability.}
\label{fig_1}
\end{figure}

The existing modeling approaches are well-posed to infer certain characteristics of human behaviors. Nevertheless, none of them have addressed the aspect of human nature that both human's \textit{task objective} \cite{My_DASC,IOC_incomplete} and \textit{variability} \cite{Variability} determine the observed behavior. The task objective reflects the intent or desire of the human and it dominates the human's behavior for the given task. The task objective is consistent over multiple demonstrations for the given task and it provides the interpretations of the human demonstrations. Meanwhile, the variability, defined as the uncertainty of human behavior, introduces stochasticity in the observed demonstrations. The variability would be inconsistent even for the same task, but its pattern can be learned from multiple demonstrations. The learned variability can be facilitated to improve the human behavior model.

In this paper, we propose a novel human behavior modeling method to address task objective and variability simultaneously. We provide an illustrative quadrotor remote control example.
In this example, the task is to land a quadrotor safely on the designated landing pad with an appropriate final position, velocity, and attitude. In Fig. \ref{fig_1}, exemplar human demonstrations are shown to clarify the motivation; the task objective can be inferred from the mean human behavior. The trajectory distribution is affected by the variability which typically makes differences from the mean human behavior for every trial, even for the same task. This is not due to an external disturbance, but is an inherent characteristic of human motor motion, which has been observed in many applications \cite{Novice_Demonstrations}, human factor \cite{Variability}, psychology  \cite{Slifkin1998}, and ergonomics \cite{Muhs2018}. It is known that variability in human performance may include complex behaviors that cannot be modeled as a simple white noise \cite{Karwowski2019}.

The proposed method aims to provide a key prerequisite condition (i.e., precise human behavior prediction) for an effective human-automation interactive control such as physical human-robot collaboration (pHRC) and shared control. In industrial pHRC, predicting a human worker's trajectory is critical to prevent collision between the human and robots. Robots can utilize the modeled human behavior for their trajectory planning and task scheduling \cite{Kinugawa2017,Kanazawa2019}. In shared control, automation can alleviate human workload and improve system performance by arbitrating human input and automation input. However, if a human behavior model is not precise, shared control schemes could conflict with the human's intent or desire (e.g., landing a quadrotor using different control strategies) \cite{My_SMC,Policy_Blending,Review_Shared_Control}. Thus, human behavior modeling is also essential for shared control schemes.

Our contributions are given as follows.
We experimentally demonstrated that the proposed method has three noticeable advantages over the existing methods. First, the proposed method can provide interpretable information about the given human demonstrations by inferring the task objective. Second, the proposed method can also improve the prediction accuracy of human behaviors for a future time-horizon and provide a confidence level of that prediction. Lastly, the proposed method is data-efficient, thus it can accurately predict future human behaviors even with a small number of human demonstrations or training data.

The rest of the paper is organized as follows. In Section \ref{Formulation}, we formulate the problem and propose a human behavior modeling method. Detailed parameter identification methods are presented in Section \ref{Modeling}. Illustrative human-subject experiment results are presented in Section \ref{Experiment}. In Section \ref{Conlusion}, conclusions are drawn.

\section{Problem Formulation} \label{Formulation}
We model human behaviors as a combination of the task objective and the variability so that human behaviors in a new, unseen situation can be accurately predicted. Mean behavior over multiple human demonstrations can be represented as the task objective which denotes the intent of a human for a given task. Variation of human behaviors, which is caused by an inherent uncertainty of the human motor motion, can be modeled as the variability. 

In the rest of this paper, the discrete-time linear time-invariant (LTI) model is used to represent a system operated by a human, similar to the other practical systems modeled as the LTI plant \cite{Linear_Model,ILQR}.
\begin{equation} \label{eq:system_dynamics}
    \mathbf{x}_{k+1} = \mathbf{A} \mathbf{x}_{k} + \mathbf{B} \mathbf{u}_{k}, \quad \mathbf{x}_0 \text{ is given}
\end{equation}
where $\mathbf{x}_k \in \mathbb{R}^n$, $\mathbf{u}_k \in \mathcal{U} \subseteq \mathbb{R}^m $, and $\mathbf{x}_0 \in \mathbb{R}^n$ denote the state of the system, control input from a human, and initial state, respectively. Note that the control input is equivalent to the human behavior in this paper.  $k$ denotes the time index and $\mathbf{A} \in \mathbb{R}^{n \times n}$ and $\mathbf{B} \in \mathbb{R}^{n \times m}$ are the system matrices. We make the following assumptions on the system dynamics so that the problem is well-posed. Note that these assumptions are common in the related literature \cite{ILQR,ILQR_systemID}.

\begin{assu} \label{assumption_stabilizable}
$(\mathbf{A},\mathbf{B})$ is known and stabilizable.
\end{assu}

\begin{assu} \label{assumption_B}
$\text{Rank}(\mathbf{B}) = m < n$.
\end{assu}

\subsection{Human Behavior Model}
The human behavior can be modeled as:
\begin{equation} \label{eq:human_input}
    \mathbf{u}_k = \bar{\mathbf{u}}_k + \mathbf{w}_k
\end{equation}
where $\bar{\mathbf{u}}_k  \in {R}^m$ denotes the \textit{task-objective-based behavior} and $\mathbf{w}_k  \in {R}^m $ denotes the variability, respectively. Note that $\bar{\mathbf{u}}_k$ is the deterministic variable and $\mathbf{w}_k$ is the stochastic variable. In the proposed method, an unknown task objective function which governs the task-objective-based behavior and a set of parameters which represents the variability are identified.

\subsubsection{Task Objective Model}
The task-objective-based behavior is assumed to minimize an unknown quadratic objective function  \cite{ILQR,My_SMC} over an infinite-horizon with a constant control gain $\mathbf{K}$:
\begin{equation} \label{eq:task_objective_based_behavior}
    \bar{\mathbf{u}}_k = \mathbf{K} \mathbf{x}_k
\end{equation}
which is a linear-quadratic regulator (LQR) gain with cost matrices $\{ \mathbf{Q},\mathbf{R},\mathbf{S} \}$. The task objective function of the infinite-horizon LQR is given as:
\begin{equation} \label{eq:LQR_cost}
    J = \sum_{k=0}^{\infty} \left( \mathbf{x}_k^T \mathbf{Q} \mathbf{x}_k + \mathbf{u}_k^T \mathbf{R} \mathbf{u}_k + 2 \mathbf{x}_k^T \mathbf{S} \mathbf{u}_k \right).
\end{equation}
Then, the control gain $\mathbf{K}$ is constrained by the followings:
\begin{equation} \label{eq:Control_Gain}
    \mathbf{K} = -(\mathbf{R} + \mathbf{B}^T\mathbf{P}\mathbf{B})^{-1} (\mathbf{B}^T \mathbf{P} \mathbf{A} + \mathbf{S}^T)
\end{equation}
\begin{equation} \label{eq:DARE}
\begin{split}
    \mathbf{P} = \mathbf{A}^T \mathbf{P} \mathbf{A} - (\mathbf{A}^T \mathbf{P} \mathbf{B} + \mathbf{S})(\mathbf{R} + \mathbf{B}^T \mathbf{P} \mathbf{B})^{-1} \\
    \times (\mathbf{B}^T \mathbf{P} \mathbf{A} + \mathbf{S}^T) + \mathbf{Q}
\end{split}
\end{equation}
\begin{equation} \label{eq:SPD}
    \text{subject to} \quad \begin{bmatrix}
    \mathbf{Q} & \mathbf{S} \\ \mathbf{S}^T & \mathbf{R}
    \end{bmatrix} \succeq 0, \quad \mathbf{R} \succ 0
\end{equation}
where $\mathbf{K} \in \mathbb{R}^{m \times n}$ denotes the unknown task-objective-based control gain; and $\mathbf{Q} \in \mathbb{R}^{n \times n}$, $\mathbf{S} \in \mathbb{R}^{n \times m}$, and $\mathbf{R} \in \mathbb{R}^{m \times m}$ are unknown task objective matrices, respectively. $\mathbf{R} \succ 0$ denotes the positive definite matrix $\mathbf{R}$. $\mathbf{P}$ is the unique semi-positive definite solution to the discrete-time algebraic Riccati equation (DARE) in \eqref{eq:DARE}.

\subsubsection{Variability Model}
The variability which denotes uncertainty of the human behavior could be a non-Gaussian distribution and state-dependent as shown in Fig. \ref{fig_1}. Thus, a mixture of normal distributions with state-dependent mean and covariance is used to model the variability. That mixture can be merged into a single state-dependent normal distribution with merged mean and covariance, by a product of the mixtures. Then, the variability can be represented by
\begin{equation} \label{eq:variability_model}
    p(\mathbf{w}_k \vert \mathbf{x}_k) \sim \mathcal{N} \left( \boldsymbol{\mu}_k (\mathbf{x}_k), \boldsymbol{\Sigma}_k (\mathbf{x}_k) \right)
\end{equation}
where $\boldsymbol{\mu}_k(\cdot)$ and $\boldsymbol{\Sigma}_k(\cdot)$ are the unknown mean and covariance, respectively. Since $\{ \boldsymbol{\mu}_k(\cdot), \boldsymbol{\Sigma}_k(\cdot) \}$ depends on the state and time index $k$, it can deal with time-varying and state-dependent variability. The covariance provides a confidence level of of the inferred variability.

\subsection{Problem Statement}
We propose a human behavior modeling method to predict the future behavior accurately. The proposed method also provides a confidence level of that prediction; e.g., predicting $\mathbf{u}_{k+1}$ and its covariance at time step $k$ in a new situation, by exploiting the given human's multiple demonstrations in various situations $ \{ \mathbf{x}_k^j, \mathbf{u}_k^j \}_{k=1, j=1}^{N_j, M}$ where $M$ denotes the number of human demonstrations and $N_j$ denotes the length of each demonstration, respectively. For each demonstration, the initial state $\mathbf{x}_0^j$ may be different. The proposed approach identifies the unknown parameters $ \{ \mathbf{Q}, \mathbf{R}, \mathbf{S} \}$ and $\{ \boldsymbol{\mu}_k(\cdot), \boldsymbol{\Sigma}_k(\cdot) \}$ separately; an identified set of matrices $ \{ \mathbf{Q}, \mathbf{R}, \mathbf{S} \}$ denotes the task objective of the human and an estimated set of state-dependent mean and covariance $\{ \boldsymbol{\mu}_k(\cdot), \boldsymbol{\Sigma}_k(\cdot) \}$ represents the variability.


\section{Parameter Identification Methods} \label{Modeling}

\begin{figure}[!b]
\centering
\includegraphics[width=0.25\textwidth]{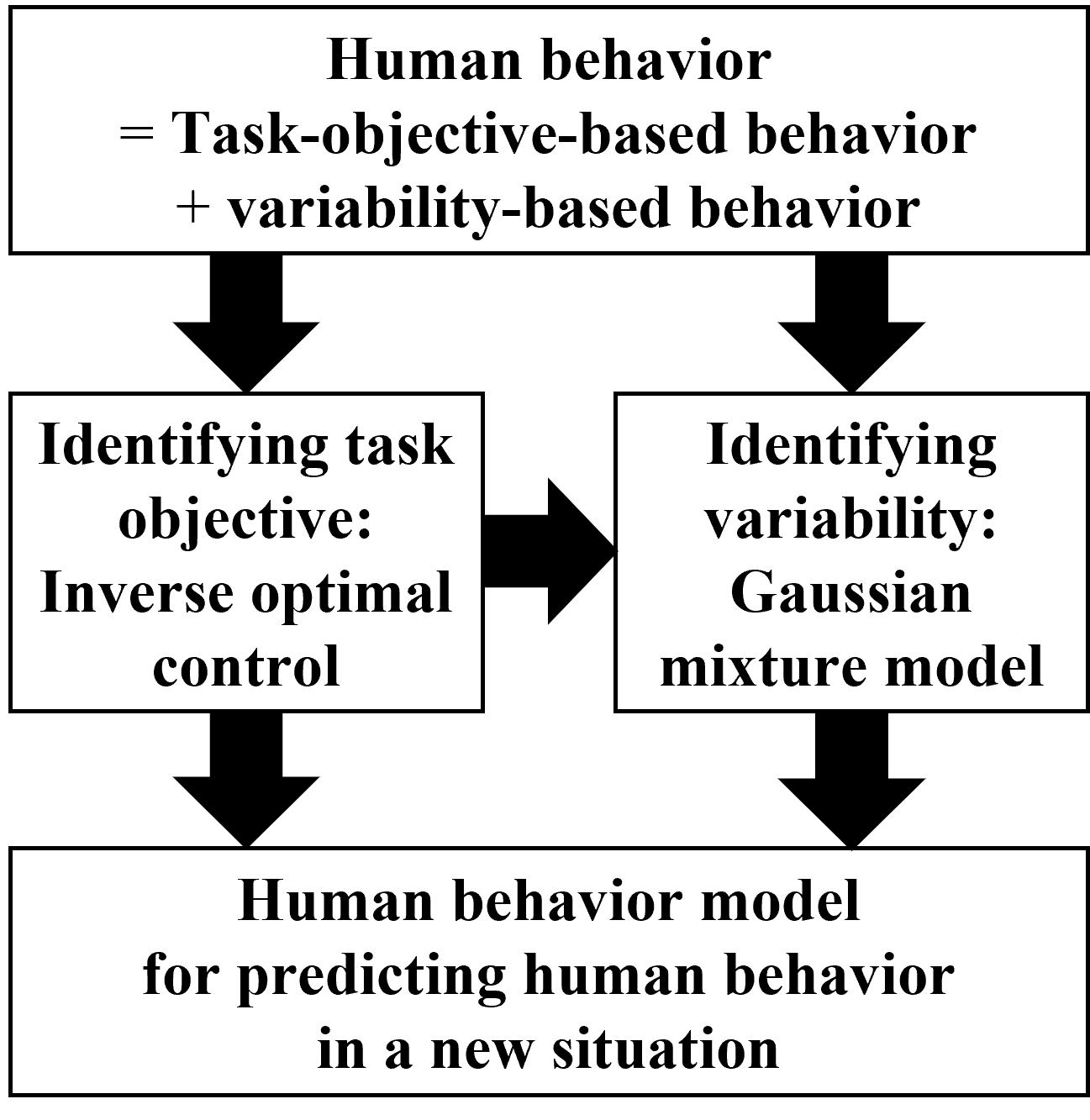}
\caption{A block scheme of the proposed method.}
\label{fig_Block}
\end{figure}

In this section, methods are presented in details to identify the parameters for the proposed human behavior model. Fig. \ref{fig_Block} presents a block scheme of the proposed method.

\subsection{Inverse Optimal Control} \label{IOC_method}
In many human-automation interactive frameworks, the IOC has been widely used to model the human behavior \cite{IOC_IRL_Review,My_DASC,My_SMC,ILQR,IOC_Jumping,IOC_Squatting,IRL,Human_Objective_Learning}. The IOC assumes that a human behaves based on their task objective, which represents a performance measure to be minimized by the human. The task objective is usually unknown. Accordingly, the IOC is employed to identify the implicit task objective as a form of an objective function, from the human behaviors interacting with the known system dynamics model. The problem to be addressed by the IOC approach for modeling human behavior is given as follows.

\begin{prob} \label{problem_IOC}
From the given state of human demonstrations $\{ \mathbf{x}_k^j  \}_{k=1,j=1}^{N_j,M}$, identify a set of estimates $\hat{\mathbf{K}}$ and $\{ \hat{\mathbf{Q}}, \hat{\mathbf{R}}, \hat{\mathbf{S}} \}$ where $\hat{\mathbf{K}}$ denotes the estimate of the task-objective-based control gain in  \eqref{eq:task_objective_based_behavior}, which is constrained by the followings:
\begin{equation} \label{eq:K_hat2}
    \hat{\mathbf{K}} = -(\hat{\mathbf{R}} + \mathbf{B}^T \hat{\mathbf{P}} \mathbf{B})^{-1}(\mathbf{B}^T \hat{\mathbf{P}} \mathbf{A} + \hat{\mathbf{S}}^T)
\end{equation}
\begin{equation} \label{eq:P_hat}
\begin{split}
    \hat{\mathbf{P}} = \mathbf{A}^T \hat{\mathbf{P}} \mathbf{A} - (\mathbf{A}^T \hat{\mathbf{P}} \mathbf{B} + \hat{\mathbf{S}}) (\hat{\mathbf{R}} + \mathbf{B}^T \hat{\mathbf{P}} \mathbf{B})^{-1} \\
    \times (\mathbf{B}^T \hat{\mathbf{P}} \mathbf{A} + \hat{\mathbf{S}}^T) + \hat{\mathbf{Q}}
\end{split}
\end{equation}

\begin{equation} \label{eq:SPD_hat}
    \begin{bmatrix}
    \hat{\mathbf{Q}} & \hat{\mathbf{S}} \\ \hat{\mathbf{S}}^T & \hat{\mathbf{R}}
        \end{bmatrix} \succeq 0, \quad \hat{\mathbf{R}} \succ 0, \quad \hat{\mathbf{P}} \succeq 0
\end{equation}
where $\{ \hat{\mathbf{Q}},\hat{\mathbf{R}},\hat{\mathbf{S}} \}$ denotes the estimate of the unknown task objective in \eqref{eq:Control_Gain}-\eqref{eq:SPD}. The set of estimates $\{ \hat{\mathbf{K}},\hat{\mathbf{Q}},\hat{\mathbf{R}},\hat{\mathbf{S}} \}$ represents the task objective model of the demonstrated human behavior. $\hat{\mathbf{P}}$ denotes the solution of the DARE.
\end{prob}
To guarantee the feasibility of Problem \ref{problem_IOC}, an assumption is given as follows. Note that Assumption \ref{assumption_K} is automatically satisfied if the human is capable of controlling the system properly \cite{My_SMC}.
\begin{assu} \label{assumption_K}
$\hat{\mathbf{K}}$ is a stabilizing control gain. Equivalently, $\lvert \rho(\mathbf{A}+\mathbf{B}\hat{\mathbf{K}}) \rvert < 1$ where $\rho(\cdot)$ denotes the spectral radius.
\end{assu}

To obtain $\hat{\mathbf{K}}$ from the given human demonstrations, we employ the least-square method \cite{DMD}:
\begin{equation} \label{eq:DMD_argue}
    \tilde{\mathbf{A}} = \mathbf{X}'\mathbf{X}^{\dagger} = \arg \min_{\tilde{\mathbf{A}}} \lVert \mathbf{X}' - \tilde{\mathbf{A}} \mathbf{X} \rVert_2
\end{equation}
where
\begin{equation} \label{eq:DMD_variables}
    \begin{gathered}
        \mathbf{X} \triangleq [ \mathbf{x}_1^1, \cdots, \mathbf{x}_{N_1-1}^1,\mathbf{x}_1^2,\cdots,\mathbf{x}_{N_2-1}^2,\cdots,\mathbf{x}_{N_M-1}^M] \\
        \mathbf{X}' \triangleq [ \mathbf{x}_2^1, \cdots, \mathbf{x}_{N_1}^1,\mathbf{x}_2^2,\cdots,\mathbf{x}_{N_2}^2,\cdots,\mathbf{x}_{N_M}^M]
    \end{gathered}
\end{equation}
and $\dagger$ denotes the pseudo-inverse. From Assumption \ref{assumption_B}, $\mathbf{B}^{\dagger} \mathbf{B} = \mathbf{I}_m$ where $\mathbf{I}_m$ denotes the $m \times m$ identity matrix. Thus, the estimate $\hat{\mathbf{K}}$ is determined by
\begin{equation} \label{eq:K_hat_computation}
    \hat{\mathbf{K}} = \mathbf{B}^{\dagger} (\tilde{\mathbf{A}} - \mathbf{A})
\end{equation}
which represents the task-objective-based behavior in the action-state-level abstraction. In the next step, $\{ \hat{\mathbf{Q}},\hat{\mathbf{R}},\hat{\mathbf{S}} \}$ is determined to represent the task-objective-based behavior in the task-objective-level abstraction. A convex optimization problem with linear matrix inequality (LMI) constraints is defined as follows to compute $\{ \hat{\mathbf{Q}},\hat{\mathbf{R}},\hat{\mathbf{S}} \}$ \cite{ILQR}.
\begin{equation} \label{eq:convex_optimization}
    \{ \hat{\mathbf{Q}},\hat{\mathbf{R}},\hat{\mathbf{S}},\hat{\mathbf{P}} \} = \arg \min_{\mathbf{Q},\mathbf{R},\mathbf{S},\mathbf{P}} \alpha^2 \quad \text{such that}
\end{equation}
\begin{equation} \label{eq:LMI_constraints1}
    \mathbf{P}  \succeq 0
\end{equation}
\begin{equation} \label{eq:LMI_constraints2}
    (\mathbf{R} + \mathbf{B}^T \mathbf{P} \mathbf{B}) \hat{\mathbf{K}}  + \mathbf{B}^T \mathbf{P} \mathbf{A} + \mathbf{S}^T = 0
\end{equation}
\begin{equation} \label{eq:LMI_constraints3}
    \mathbf{A}^T \mathbf{P} \mathbf{A} - \mathbf{P} + (\mathbf{A}^T\mathbf{P} \mathbf{B} + \mathbf{S}) \hat{\mathbf{K}} + \mathbf{Q} = 0
\end{equation}
\begin{equation} \label{eq:LMI_constraints4}
    \mathbf{I}_{n+m} \preceq \begin{bmatrix}
    \mathbf{Q} & \mathbf{S} \\ \mathbf{S}^T & \mathbf{R}
    \end{bmatrix} \preceq \alpha \mathbf{I}_{n+m}
\end{equation}
where $\alpha^2$ is minimized such that a scalar ambiguity is resolved and a unique solution is found \cite{ILQR_systemID}. Since the feasibility of  \eqref{eq:convex_optimization} with constraints in \eqref{eq:LMI_constraints1}-\eqref{eq:LMI_constraints4} under Assumption \ref{assumption_K} was shown in \cite{ILQR_feasibility}, the solution of  \eqref{eq:convex_optimization} always exists and is unique.

\begin{rem}
An estimate $\hat{\mathbf{K}}$ is not necessarily the same as $\mathbf{K}$. For instance, if $\mathbf{w}_k$ is a linear function of $\mathbf{x}_k$, $\mathbf{w}_k = \mathbf{L}\mathbf{x}_k + \mathbf{v}_k$, where $\mathbf{L} \in \mathbb{R}^{m \times n}$, $\mathbf{v}_k \in \mathbb{R}^m$, and $\mathbf{v}_k \sim \mathcal{N}(0,\boldsymbol{\Sigma}_{\mathbf{v}})$, then $\hat{\mathbf{K}} = \mathbf{K} + \mathbf{L}$. However, the identified $\hat{\mathbf{K}}$ represents a dominant behavior of the demonstrated human behaviors, by solving a least-square problem in \eqref{eq:DMD_argue}, with any forms of the variability. Thus, $\hat{\mathbf{K}}$ can be considered as a feasible estimate for the task-objective-based behavior control gain.
\end{rem}

\begin{rem}
In some practical applications, $\mathbf{S}=0$ is assumed or preferred \cite{ILQR,ILQR_systemID}. However, if this is assumed, then Problem \ref{problem_IOC} may not be feasible with an arbitrary stabilizing $\mathbf{K}$ \cite{ILQR_feasibility}. The inequality in \eqref{eq:LMI_constraints4} is more conservative than  \eqref{eq:SPD}, but it provides additional accuracy in computation by preventing the solution of $\alpha$ from becoming too small.
\end{rem}

\subsection{Variability Parameter Identification} \label{Variability_method}

The variability is defined as an uncertainty in human behavior which appears in their multiple demonstrations \cite{Variability}. It is intrinsically objective-less and stochastic, so the IOC is insufficient for identifying the variability. In other applications, it has been shown that stochasticity depends on the context, such as the system states, environment, and mission objectives \cite{Human_Behavior_Uncertainty,ProMPs,TP_Tutorial}. In Fig. \ref{fig_1}, for instance, in the quadrotor landing mission, a human tends to be more consistent when the vehicle is close to the landing pad. We propose a method for learning variability from human's multiple demonstrations. The proposed scheme employs the Gaussian mixture model (GMM) to model the variability whose stochastic properties are possibly non-Gaussian and/or multi-modal, since the GMM can represent a multi-modal and non-Gaussian distribution as a mixture of multiple Gaussian distributions. Thus, the accuracy of the modeling is improved compared with a single zero-mean Gaussian approximation \cite{ProMPs}. To train the GMM, we use the expectation-maximization (EM) algorithm to find parameters which maximize the log-likelihood \cite{EM1,EM2}. The trained variability can be exploited in a new environment to reproduce the variability, so that the accuracy of predicting the human behavior can be improved. The following problem will be addressed to identify the variability.

\begin{prob} \label{problem_variability}
From the given human demonstrations $\{ \mathbf{x}_k^j, \mathbf{u}_k^j \}_{k=1,j=1}^{N_j, M}$ and an estimate $\hat{\mathbf{K}}$ from  \eqref{eq:K_hat_computation}, identify a set of parameters such that an estimate of the variability can be modeled by those parameters.
\begin{equation} \label{eq:variability_estimate}
    \hat{\mathbf{w}}_k^j \triangleq \mathbf{u}_k^j - \hat{\mathbf{K}} \mathbf{x}_k^j
\end{equation}
\begin{equation}
    p(\hat{\mathbf{w}}_k^j \vert \mathbf{x}_k^j) \sim \mathcal{N} \left( \hat{\boldsymbol{\mu}}_k(\mathbf{x}_k^j), \hat{\boldsymbol{\Sigma}}_k(\mathbf{x}_k^j) \right)
\end{equation}
where $\{ \hat{\boldsymbol{\mu}}_k(\cdot), \hat{\boldsymbol{\Sigma}}_k(\cdot) \}$ denotes the estimate of mean and covariance of the variability and the solution of Problem \ref{problem_variability}.
\end{prob} 

\subsubsection{Encoding Variability}

Probabilistic encoding methods for the continuous movements have been widely applied to robotics, imitation learning, and human motor skill modeling \cite{ProMPs,TP_Tutorial,Imitation_learning}. The existing probabilistic approaches aim to reproduce the learned movements in a new situation. We employ probabilistic movement encoding techniques in a different context to model the variability of the human behavior \cite{TP_Tutorial}.

The probabilistic encoding technique can model the conditional probability $p(\hat{\mathbf{w}}_k^j \vert \mathbf{x}_k^j)$ as a function of input and output. This function is approximated as a GMM with unknown parameters. In this formulation, the input is the state $\mathbf{x}_k^j$ and the output is the variability $\hat{\mathbf{w}}_k^j$. The superscript $j$ will be omitted in the following discussion for simplicity. Let $\mathcal{I}$ and $\mathcal{O}$ be representing the input $\boldsymbol{\xi}_k^{\mathcal{I}}$ and output $\boldsymbol{\xi}_k^{\mathcal{O}}$, respectively. At each time step $k$, the data point $\boldsymbol{\xi}_k$ is divided into the input and output. The GMM encodes this data point with a set of parameters $\{ \bar{h}^i, \bar{\boldsymbol{\mu}}^i, \bar{\boldsymbol{\Sigma}}^i \}_{i=1}^G$ where $i \in \{1,\cdots,G \}$ and $G$ denotes the number of Gaussian components.
\begin{equation} \label{eq:GMM_xi_mu}
    \boldsymbol{\xi}_k \triangleq \begin{bmatrix}
    \boldsymbol{\xi}_k^{\mathcal{I}} \\ \boldsymbol{\xi}_k^{\mathcal{O}}
    \end{bmatrix}
    = \begin{bmatrix}
    \mathbf{x}_k \\ \hat{\mathbf{w}}_k
    \end{bmatrix}, \quad
    \bar{\boldsymbol{\mu}}^i = \begin{bmatrix} \bar{\boldsymbol{\mu}}^{i,\mathcal{I}} \\ \bar{\boldsymbol{\mu}}^{i,\mathcal{O}}
    \end{bmatrix}
\end{equation}

\begin{equation} \label{eq:GMM_Sigma}
    \bar{\boldsymbol{\Sigma}}^i = \begin{bmatrix}
    \bar{\boldsymbol{\Sigma}}^{i,\mathcal{I}} & \bar{\boldsymbol{\Sigma}}^{i,\mathcal{I} \mathcal{O}} \\
    \bar{\boldsymbol{\Sigma}}^{i,\mathcal{O} \mathcal{I}} & \bar{\boldsymbol{\Sigma}}^{i,\mathcal{O}}
    \end{bmatrix}
\end{equation}
where $\bar{h}^i \in [0,1]$ denotes the priors (the probability that a data point belongs to the $i$-th Gaussian component) and $\sum_{i=1}^G \bar{h}^i = 1$. $\{ \bar{\boldsymbol{\mu}}^i, \bar{\boldsymbol{\Sigma}}^i \}_{i=1}^G$ denotes the mean and covariance of the $i$-th Gaussian component. A set of parameters $\{ \bar{h}^i, \bar{\boldsymbol{\mu}}^i, \bar{\boldsymbol{\Sigma}}^i \}_{i=1}^G$ is trained using the standard EM algorithm \cite{TP_Tutorial,EM1,EM2} from the given human demonstrations. We employ the k-means algorithm \cite{k_mean} to provide a good initial guess to the EM algorithm, since the initial guess has an impact on the performance and accuracy of the EM algorithm. The Gaussian mixture regression (GMR) relies on the estimated GMM parameters to compute the conditional probability $p(\boldsymbol{\xi}_k^{\mathcal{O}} \vert \boldsymbol{\xi}_k^{\mathcal{I}}) = p(\hat{\mathbf{w}}_k \vert \mathbf{x}_k)$, with the current state $\mathbf{x}_k = \boldsymbol{\xi}_k^{\mathcal{I}}$. At each time step $k$, the conditional probability is estimated as a linear combination of Gaussian distributions.
\begin{equation} \label{eq:GMR_probability}
   p(\hat{\mathbf{w}}_k \vert \mathbf{x}_k) \sim \sum_{i=1}^{G} h^i(\mathbf{x}_k) \mathcal{N}( \hat{\boldsymbol{\mu}}_k^i(\mathbf{x}_k), \hat{\boldsymbol{\Sigma}}^i )
\end{equation}
where
\begin{equation} \label{eq:GMR_mu_Sigma_h}
\begin{split}
    \hat{\boldsymbol{\mu}}_k^i(\mathbf{x}_k) & = \bar{\boldsymbol{\mu}}^{i,\mathcal{O}} + \bar{\boldsymbol{\Sigma}}^{i,\mathcal{O} \mathcal{I}} (\bar{\boldsymbol{\Sigma}}^{i,\mathcal{I}})^{-1} \left( \mathbf{x}_k - \bar{\boldsymbol{\mu}}^{i,\mathcal{I}} \right) \\
    \hat{\boldsymbol{\Sigma}}^i & = \bar{\boldsymbol{\Sigma}}^{i,\mathcal{O}} - \bar{\boldsymbol{\Sigma}}^{i,\mathcal{O} \mathcal{I}} (\bar{\boldsymbol{\Sigma}}^{i,\mathcal{I}})^{-1} \bar{\boldsymbol{\Sigma}}^{i,\mathcal{I} \mathcal{O}} \\
    h^i(\mathbf{x}_k) & = \frac{\bar{h}^i \mathcal{N} \left( \mathbf{x}_k \vert \bar{\boldsymbol{\mu}}^{i,\mathcal{I}}, \bar{\boldsymbol{\Sigma}}^{i,\mathcal{I}} \right)}{\sum_{g=1}^G \bar{h}^g \mathcal{N} \left( \mathbf{x}_k \vert \bar{\boldsymbol{\mu}}^{g,\mathcal{I}}, \bar{\boldsymbol{\Sigma}}^{g,\mathcal{I}} \right) }
\end{split}
\end{equation}
and $\mathcal{N}(\mathbf{x} \vert \boldsymbol{\mu}, \boldsymbol{\Sigma})$ denotes a value of the Gaussian function with input $\mathbf{x}$, mean $\boldsymbol{\mu}$, and covariance $\boldsymbol{\Sigma}$. $h^i(\cdot)$ denotes the activation weight of each Gaussian component and $\Sigma_{i=1}^G h^i(\mathbf{x}_k) = 1$. The above multi-modal distribution can be approximated as a single distribution \cite{TP_Tutorial}.
\begin{equation} \label{eq:GMR_single}
    p(\hat{\mathbf{w}}_k \vert \mathbf{x}_k) \sim \mathcal{N} \left( \hat{\boldsymbol{\mu}}_k(\mathbf{x}_k), \hat{\boldsymbol{\Sigma}}_k(\mathbf{x}_k) \right)
\end{equation}
where
\begin{equation} \label{eq:GMR_single_mu_sigma}
\begin{split}
    \hat{\boldsymbol{\mu}}_k (\mathbf{x}_k) & = \sum_{i=1}^G h^i(\mathbf{x}_k) \hat{\boldsymbol{\mu}}_k^i (\mathbf{x}_k) \\
    \hat{\boldsymbol{\Sigma}}_k (\mathbf{x}_k) = \sum_{i=1}^G & h^i(\mathbf{x}_k) \left( \hat{\boldsymbol{\Sigma}}^i + \hat{\boldsymbol{\mu}}_k^i (\mathbf{x}_k) \hat{\boldsymbol{\mu}}_k^i (\mathbf{x}_k)^T \right) \\
    - & \hat{\boldsymbol{\mu}}_k (\mathbf{x}_k) \hat{\boldsymbol{\mu}}_k (\mathbf{x}_k)^T.
\end{split}
\end{equation}

\begin{rem}
The GMM parameter estimation requires a relatively large amount of computation because the EM algorithm requires iterative computation. However, the GMR with a newly observed data point $\mathbf{x}_{k+1}$ at time step $k+1$ only requires simple computation to obtain $ \{ \hat{\boldsymbol{\mu}}^{k+1} (\mathbf{x}_{k+1}), \hat{\boldsymbol{\Sigma}}_{k+1} (\mathbf{x}_{k+1}) \}$, once the GMM parameter is stored in memory. $\hat{\boldsymbol{\Sigma}}^i$ and $\bar{\boldsymbol{\Sigma}}^{i,\mathcal{O} \mathcal{I}} (\bar{\boldsymbol{\Sigma}}^{i,\mathcal{I}})^{-1}$ can be computed offline to reduce the computational load when $\hat{\boldsymbol{\mu}}_k^i (\mathbf{x}_k)$ and $h^i(\mathbf{x}_k)$ are being updated. In this regard, the proposed method can update the variability online \cite{TP_Tutorial}. The merged distribution in \eqref{eq:GMR_single} is preferred to represent the distribution in a simpler form. Note that the multi-modal distribution in \eqref{eq:GMR_probability} can be used when that mode information is necessary.
\end{rem}

\subsubsection{Task-Parameterized Variability}
A task parameter refers to the variable which encodes environment, context, or situation of each demonstration, such as the initial position and target position \cite{TP_Tutorial,TP_Minimal_Intervention1,TP_Minimal_Intervention2}. The learned GMM model can be employed in different situations by simply changing the task parameter. Thus, the task parameter has an important role when the proposed human behavior modeling reproduces the trained variability in a new situation.

In the proposed modeling scheme, a set of task parameters represents the coordinate system which is used to observe the human demonstrations. Each coordinate system is defined as a set of linear transformation matrices $\mathbf{T}^p$ and bias (or origin) $\mathbf{b}^p$ of the observer with $p\in \{ 1,\cdots,P \}$ different coordinate systems. $\mathbf{Z}_k^p \in \mathbb{R}^{n+m}$ denotes the demonstrated human behavior at time step $k$, which is observed in the $p$-th coordinate perspective, and it is represented as:
\begin{equation} \label{eq:state_frame}
    \mathbf{Z}_k^p = (\mathbf{T}^p)^{-1} (\boldsymbol{\xi}_k - \mathbf{b}^p)
\end{equation}
where
\begin{equation} \label{eq:T_b}
\begin{gathered}
    \mathbf{T}^p \triangleq \begin{bmatrix}
    \mathbf{T}^{p,\mathcal{I}} & 0 \\ 0 & \mathbf{T}^{p,\mathcal{O}}
    \end{bmatrix} \in \mathbb{R}^{(n+m) \times (n+m)} \\
    \mathbf{b}^p \triangleq \begin{bmatrix}
    \mathbf{b}^{p,\mathcal{I}} \\ \mathbf{b}^{p,\mathcal{O}}
    \end{bmatrix} \in \mathbb{R}^{n+m}
\end{gathered}
\end{equation}
and the GMM is trained in $P$ different perspectives. The GMM parameter is a set of $\{ \bar{h}^i, \{ \bar{\boldsymbol{\mu}}^{i,p}, \bar{\boldsymbol{\Sigma}}^{i,p} \}_{p=1}^P \}_{i=1}^G$, which is learned using the EM algorithm. The learned task-parameterized GMM model is merged into a single Gaussian distribution to be used for reproducing the variability with a new task parameter.
\begin{equation} \label{eq:TP_GMR}
    \begin{split}
        \bar{\boldsymbol{\Sigma}}^i = & \left( \sum_{p=1}^P \left( \mathbf{T}^p  \bar{\boldsymbol{\Sigma}}^{i,p}  (\mathbf{T}^p)^T \right)^{-1} \right)^{-1} \\
        \bar{\boldsymbol{\mu}}^i = \bar{\boldsymbol{\Sigma}}^i & \sum_{p=1}^P \left( \mathbf{T}^p  \bar{\boldsymbol{\Sigma}}^{i,p}  (\mathbf{T}^p)^T \right)^{-1} \left( \mathbf{T}^p \bar{\boldsymbol{\mu}}^{i,p} + \mathbf{b}^p \right)
    \end{split}
\end{equation}
and the merged Gaussian model with $\{ \bar{h}^i, \bar{\boldsymbol{\mu}}^i, \bar{\boldsymbol{\Sigma}}^i \}_{i=1}^G$, which encodes all information in $P$ different perspectives, is used for the GMR to provide more generality for a new situation. Comprehensive details, including inferring the task objective and identifying the variability, are given in Algorithm \ref{algorithm1}.


\begin{algorithm} 
 \caption{The proposed human behavior modeling.}
 \begin{algorithmic}[1] \label{algorithm1}
 \renewcommand{\algorithmicrequire}{\textbf{Input:}}
 \renewcommand{\algorithmicensure}{\textbf{Output:}}
 \REQUIRE System matrices $(\mathbf{A},\mathbf{B})$ and human demonstrations (training data) $\{ \mathbf{x}_k^j, \mathbf{u}_k^j \}_{k=1,j=1}^{N_j, M}$.
 \ENSURE  Inferred task objective $\{ \hat{\mathbf{Q}}, \hat{\mathbf{R}}, \hat{\mathbf{S}} \}$ and identified variability parameter $\{ \hat{\boldsymbol{\mu}}_k(\cdot), \hat{\boldsymbol{\Sigma}}_k(\cdot) \}$.
  \STATE Solve the least-square problem in  \eqref{eq:DMD_argue}-\eqref{eq:K_hat_computation} to obtain $\hat{\mathbf{K}}$ from the given human demonstrations $\{ \mathbf{x}^j_k, \mathbf{u}^j_k \}_{k=1,j=1}^{N_j, M}$.
  \STATE Solve the convex optimization problem in  \eqref{eq:convex_optimization}-\eqref{eq:LMI_constraints4} for inferring the task objective $\{ \hat{\mathbf{Q}}, \hat{\mathbf{R}}, \hat{\mathbf{S}} \}$.
   \FOR{$j \in \{1,\cdots,M \}$}
    \FOR{$k\in\{1,\cdots,N_j\}$} 
      \STATE{Generate $\hat{\mathbf{w}}_k^j = \mathbf{u}_k^j - \hat{\mathbf{K}}\mathbf{x}_k^j$  in \eqref{eq:variability_estimate}.} 
      \ENDFOR
    \ENDFOR
 \STATE Generate $\boldsymbol{\xi}_k^j = [(\mathbf{x}_k^j)^T \quad (\hat{\mathbf{w}}_k^j)^T]^T$.
 \FOR{$p \in \{1,\cdots,P \}$}
 \STATE{Generate the task-parameterized trajectory $\mathbf{Z}_k^{j,p} = (\mathbf{T}^p)^{-1}(\boldsymbol{\xi}_k^j - \mathbf{b}^p)$ as  \eqref{eq:state_frame}.}
 \STATE{$\mathbf{Z}^p = [\mathbf{Z}_1^{1,p},\cdots,\mathbf{Z}_{N_1}^{1,p},\mathbf{Z}_1^{2,p},\cdots,\mathbf{Z}_{N_2}^{2,p},\cdots,\mathbf{Z}_{N_M}^{M,p}]$}
 \STATE{Obtain $\{ \bar{h}^i, \bar{\boldsymbol{\mu}}^{i,p}, \bar{\boldsymbol{\Sigma}}^{i,p} \}_{i=1}^G$ by training the GMM with the EM algorithm and $G$ Gaussian components.}
 \ENDFOR
 \STATE{Merge $P$ different Gaussian components into a single distribution $\{ \bar{h}^i, \bar{\boldsymbol{\mu}}^i, \bar{\boldsymbol{\Sigma}}^i \}_{i=1}^G$ with $i=\{1,\cdots,G \}$ as  \eqref{eq:TP_GMR}.}
 \FOR{$i \in \{1,\cdots,G \}$}
 \STATE{Compute $\{ \hat{\boldsymbol{\mu}}_k^i(\cdot), \hat{\boldsymbol{\Sigma}}^i \}$ and $h^i(\cdot)$} using  \eqref{eq:GMR_single_mu_sigma}.
 \ENDFOR
 \STATE{Merge $G$ different Gaussian components into a single distribution $\{ \hat{\boldsymbol{\mu}}_k(\cdot), \hat{\boldsymbol{\Sigma}}_k(\cdot) \}$ as  \eqref{eq:GMR_single_mu_sigma}}.
 \RETURN $\{ \hat{\mathbf{Q}}, \hat{\mathbf{R}}, \hat{\mathbf{S}} \}$ and $\{ \hat{\boldsymbol{\mu}}_k(\cdot), \hat{\boldsymbol{\Sigma}}_k(\cdot) \}$.
 \end{algorithmic}
 \end{algorithm}



\section{Human-Subject Experiment} \label{Experiment}
We demonstrate the proposed human behavior modeling scheme using an illustrative quadrotor landing example. This human-subject experimental study is approved by the Institutional Review Board at Purdue University (protocol number: IRB-2020-755). A single-subject case study and a multiple-subject case study were conducted.

\subsection{Testbed}
\begin{figure}[!tb]
\centering
\includegraphics[width=0.48\textwidth]{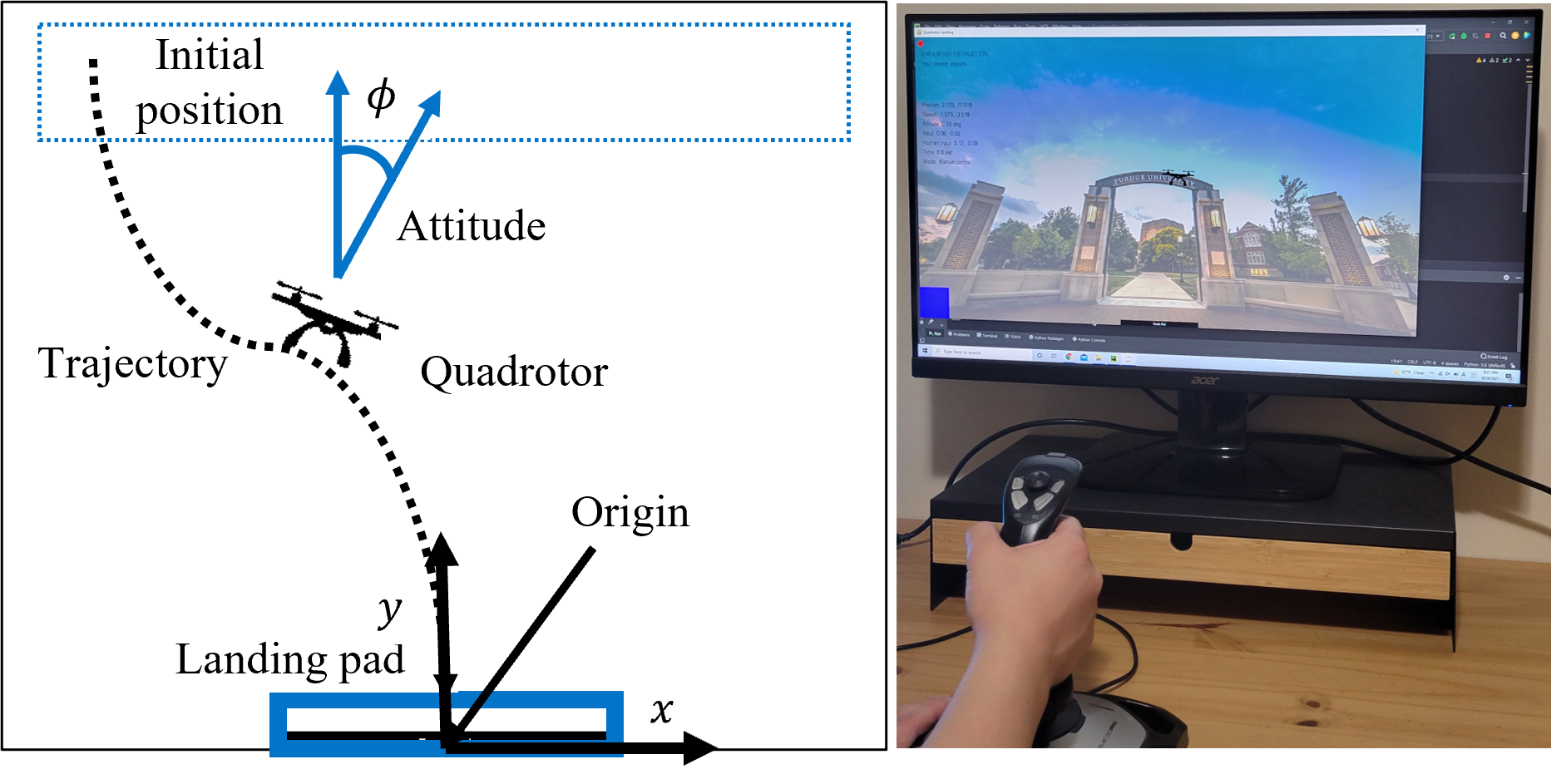}
\caption{(Left) Schematic diagram of the quadrotor landing simulator. (Right) Physical configuration of the testbed with a human operator.}
\label{fig_2}
\end{figure}

A 3-DOF quadrotor landing simulator has been developed as a testbed \cite{My_DASC,My_SMC} to conduct simulations and human-subject experiments (Fig. \ref{fig_2}). A human operator is requested to land a quadrotor on the landing pad using a joystick by controlling the quadrotor's angular acceleration and thrust. Visual feedbacks are given to the human operator via a monitor. The discrete-time linearized quadrotor dynamics is adopted with the state vector $\mathbf{x}_k = [x_k, y_k, \phi_k,\dot{x}_k,\dot{y}_k,\dot{\phi}_k]^T$, which consists of the position $(x,y)$, attitude ($\phi$), velocity $(\dot{x},\dot{y})$, and angular velocity $(\dot{\phi})$ of the quadrotor. The linearized system dynamics with respect to an equilibrium point \cite{Quadrotor_Model,My_SMC} is given by
\begin{equation} \label{eq:simulator_dynamics_A}
    \mathbf{A} = \mathbf{I}_6 + \Delta t 
    \begin{bmatrix}
    0 & 0 & 0 & 1 & 0 & 0 \\
    0 & 0 & 0 & 0 & 1 & 0 \\
    0 & 0 & 0 & 0 & 0 & 1 \\
    0 & 0 & g & 0 & 0 & 0 \\
    0 & 0 & 0 & 0 & k_1 & 0 \\
    0 & 0 & k_2 & 0 & 0 & k_3
    \end{bmatrix}
\end{equation}
\begin{equation} \label{eq:simulator_dynamics_B}
    \mathbf{B} = \Delta t \begin{bmatrix}
    0 & 0 \\
    0 & 0 \\
    0 & 0 \\
    0 & 0 \\
    0 & 1/m \\
    1/I_{x} & 0
    \end{bmatrix}
\end{equation}
where $\Delta t = 0.05$ seconds denotes the time interval for discretization, $g = 9.8 \ m/s^2$ denotes the gravitational acceleration, $\{ k_1, k_2, k_3 \} = \{ -0.1, -1, -30 \}$ is a set of controller parameters to stabilize the quadrotor, $m=0.25 \ kg$ is the mass of the quadrotor, and $I_x = 0.01 \ kg \cdot m^2$ denotes the moment of inertia with respect to the rotational axis, respectively. The control input is $\mathbf{u}_k = [ u_{1,k}, u_{2,k}]^T \in [-1, 1]^2$ where $u_{1,k}$ and $u_{2,k}$ denote the angular acceleration and thrust, respectively. The position domain $\mathcal{X} = [-3,3] \times [0,3.5]$ in meter is fixed, and the initial position of the quadrotor is randomly generated and uniformly distributed in $x_0 \in [-2,2]$ and $y_0 \in [2.5,3]$, respectively. All other initial states are set to zero. The mission objective is to land the quadrotor with an appropriate final speed ($< 0.1 \ m/s $) and final attitude ($< 5^\circ$) on the landing pad.

\subsection{Single-Subject Case Study} 
A single-subject case study was
conducted to demonstrate the proposed human behavior modeling approach. A human-subject, who successfully landed the quadrotor more than 200 times in our pilot study, participated in the experiment.
The goal of this case study is to present detailed results of the proposed method. We will show how the proposed method provides explanatory factors for human behavior and can accurately predict human behavior through variability identification.

\subsubsection{Procedure}

Two different strategy-level objectives are given to the human operator to test that the proposed modeling approach can reveal and explain the difference between them.
\begin{itemize}
    \item \textit{Control Strategy 1} (CS1): reduce the horizontal distance from the origin first, and then go down to the landing pad.
    \item \textit{Control Strategy 2} (CS2): move in a straight line to the landing pad, while minimizing the attitude control.
\end{itemize}

\begin{figure}[!tb]
\centering
\includegraphics[width=0.48\textwidth]{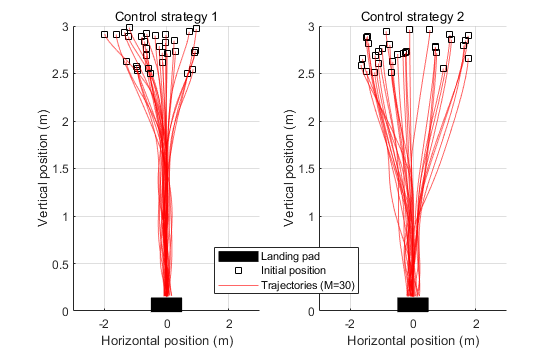}
\caption{Collected trajectories under (Left) CS1 and (Right) CS2.}
\label{fig_5}
\end{figure}

Two types of data sets were obtained. First, the human-subject conducted $30$ trials for CS1, and then $30$ trials for CS2, as shown in Fig. \ref{fig_5} (total of $60$ trials). One-minute break was given between recording CS1 and CS2. This data set is used as \textit{training data} for the human model using the three modeling methods (Algorithm \ref{algorithm1}, IOC-only, and GMR-only). Second, after another one-minute break, the human-subject conducted $3$ trials for CS1, and then $3$ trials for CS2 (total $6$ additional trials). One minute break was given between CS1 and CS2. The second data set is used as \textit{testing data}, i.e., the modeling methods are employed to predict the future trajectory of the second data set. Only the initial condition of the testing data is provided to the modeling methods. Then, the modeling methods can predict the future trajectory using the trained human model. The testing data is regarded as the \textit{ground truth} to validate that the predicted trajectory is accurate.

The number of Gaussian components is set to $G=5$ and the task parameters
\begin{equation} \label{eq:simulation_TP_b}
    \mathbf{b}^1 = \begin{bmatrix}
    \mathbf{x}_0^T & 0 & 0
    \end{bmatrix}^T, \quad
    \mathbf{b}^2 = \begin{bmatrix}
    \mathbf{x}_N^T & 0 & 0
    \end{bmatrix}^T
\end{equation}
\begin{equation} \label{eq:simulation_TP_T_phi}
    \mathbf{T}^1 = \mathbf{I}_2, \quad
    \mathbf{T}^2 = \begin{bmatrix}
    \cos(\phi_N) & -\sin(\phi_N) \\ \sin(\phi_N) & \cos(\phi_N)
    \end{bmatrix}
\end{equation}
are used to train the proposed human model.

\subsubsection{Task Objective Inference} \label{Task_Objective_Inference}
\begin{figure}[!tb]
\centering
\includegraphics[width=0.48\textwidth]{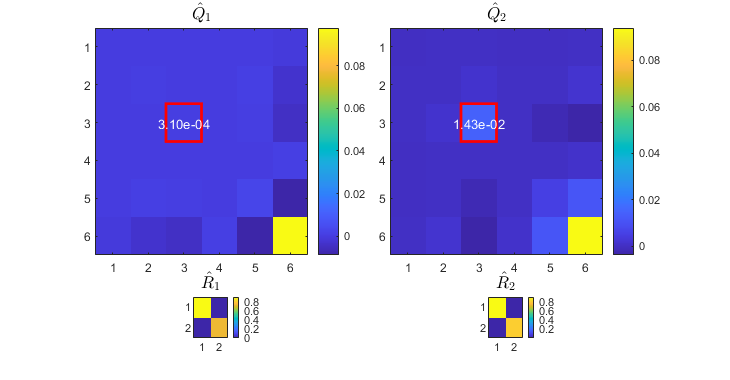}
\caption{An inferred task objective $\{ \hat{\mathbf{Q}},\hat{\mathbf{R}} \}_s, s \in \{1,2\}$ with normalization. The red-box represents the third diagonal element of $\hat{\mathbf{Q}}_s$ (a quadratic cost element on the attitude $\phi$).}
\label{fig_6}
\end{figure}

For each control strategy, an estimate $\hat{\mathbf{K}}_s$ and a corresponding task objective $\{ \hat{\mathbf{Q}},\hat{\mathbf{R}},\hat{\mathbf{S}} \}_s$ were obtained using the IOC technique in  \eqref{eq:convex_optimization}-\eqref{eq:LMI_constraints4} where $s= \{1, 2 \}$ denotes the index of each control strategy. $\hat{\mathbf{K}}_1$ and $\hat{\mathbf{K}}_2$ are obviously different due to the discrepancy between the two control strategies ($\lVert \hat{\mathbf{K}}_1 - \hat{\mathbf{K}}_2 \rVert_F = 0.1254$), but this is not interpretable by itself. On the other hand, the inferred task objective provides more information. In Fig. \ref{fig_6}, two inferred task objective sets, $\{ \hat{\mathbf{Q}}, \hat{\mathbf{R}} \}_1$ and $\{ \hat{\mathbf{Q}}, \hat{\mathbf{R}} \}_2$, are visualized. Note that these task objective matrices are normalized by dividing them with the maximum eigenvalues of each augmented square matrix (in the form of  \eqref{eq:SPD_hat}). One noticeable point is that the difference between the third diagonal element of $\hat{\mathbf{Q}}_1$ and $\hat{\mathbf{Q}}_2$ is much larger than that between the other diagonal elements (e.g., $\hat{\mathbf{Q}}_2(3,3)$ is about 46 times larger than $\hat{\mathbf{Q}}_1(3,3)$). It means CS2 is much conservative in the attitude maneuver, compared with CS1, since these elements represent the quadratic cost on the attitude. Thus, the inferred task objective matrices can provide explainable properties of human behavior, which may not be available when only control strategies are compared. The inferred task objective matrices can also be used as a performance measure in human-automation interactive control schemes \cite{My_SMC}.

\subsubsection{Variability Identification} \label{Variability_Identification}
\begin{figure}[!tb]
\centering
\includegraphics[width=0.48\textwidth]{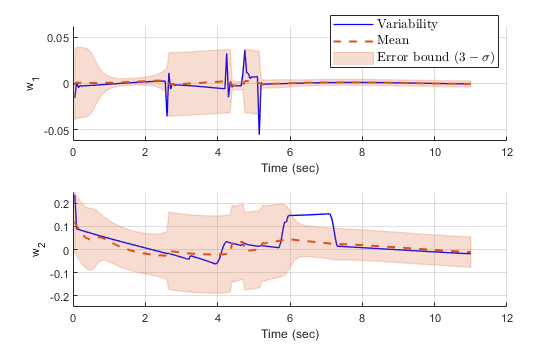}
\caption{The variability from a testing data set (blue solid-line). The inferred variability mean (red dotted-line) and covariance (red area) from the training data set and current state. The testing data and training data from CS1 are used.}
\label{fig_8}
\end{figure}

The estimates of the variability in  \eqref{eq:variability_estimate} for the human demonstrations are shown in Fig. \ref{fig_8}. The corresponding mean and covariance from the GMR in \eqref{eq:GMR_single_mu_sigma} are shown. Note that the mean and covariance of the GMR $\{ \hat{\boldsymbol{\mu}}_k(\mathbf{x}_k), \hat{\boldsymbol{\Sigma}}_k(\mathbf{x}_k) \}$ were computed using the system dynamics  \eqref{eq:system_dynamics}, the testing data at time index $k-1$, and the training data. Thus, the GMR mean and covariance are used to predict the variability a single step ahead. For each testing data set, the variability is properly bounded by the GMR $3-\sigma$ error bound.
The proposed modeling approach can successfully identify the variability. For all testing data, $97.8 \%$ and $95.2 \%$ of the variability $\hat{\mathbf{w}}_k$ are within the $3-\sigma$ error bound in average, for each axis ($w_1$ and $w_2$). In Fig. \ref{fig_8}, the GMR error bound of $w_1$ was adjusted in response to the sudden changes in the variability at times around $2.5 - 5.0 \text{ sec}$.

\subsubsection{Trajectory Prediction}\label{comparison_trajectory}
A comparison study was conducted to compare the trajectory prediction accuracy of three methods: the proposed modeling method, the IOC-only method, and the GMR-only method. The trajectory prediction is widely used to design human-automation interactive control schemes such as shared control \cite{Review_Shared_Control,Topology,My_SMC}. Thus, the trajectory prediction for a finite time-horizon was examined.

In this comparison study, the quadrotor trajectories in three seconds future time-horizon for the current states are predicted by each method. The three-second horizon (or 60 steps in the discretize-time system since $\Delta t = 0.05$ seconds) is about $25\%$ portion of each entire trial since the average landing time was about $12$ seconds.
For each modeling method, the predicted human control input $\hat{\mathbf{u}}_k$ for $k \in [0,59]$ is computed for a given initial state $\mathbf{x}_0 = \hat{\mathbf{x}}_0$. Since the system dynamics is given in \eqref{eq:system_dynamics}, we can propagate the predicted trajectories $\hat{\mathbf{x}}_k = \mathbf{A} \hat{\mathbf{x}}_{k-1} + \mathbf{B} \hat{\mathbf{u}}_{k-1}$ for $k \in [1, 60]$ using the predicted human control input. In the proposed method, the control input is predicted as $\hat{\mathbf{u}}_k = \hat{\mathbf{K}}\hat{\mathbf{x}}_k + \hat{\boldsymbol{\mu}}_k(\hat{\mathbf{x}}_k) $. In the IOC-only method, the predicted human input is given by a feedback control form $\hat{\mathbf{u}}_k = \bar{\mathbf{u}}_k = \hat{\mathbf{K}}\hat{\mathbf{x}}_k$.
The GMR-only method identifies the mean and covariance of $p(\hat{\mathbf{u}}_k \vert \hat{\mathbf{x}}_k)$ directly from the given human demonstrations. The methodology applied here is the same as  \eqref{eq:GMM_xi_mu}-\eqref{eq:TP_GMR}, except that $\hat{\mathbf{w}}_k$ is replaced by $\hat{\mathbf{u}}_k$. See Appendix for the details.

To investigate the prediction accuracy, we employed the testing data. Let $\tilde{\mathbf{x}}_k$ for $k\in[0,60]$ be the recorded quadrotor trajectory in the testing data. Only the initial state $\tilde{\mathbf{x}}_0$ is provided to each modeling method, i.e., $\hat{\mathbf{x}}_0 = \tilde{\mathbf{x}}_0$ to predict the future trajectory. Then, we use the quadrotor trajectory in the testing data as the \textit{ground truth} to compute the root mean square error (RMSE).
\begin{equation} \label{eq:RMSE}
\begin{gathered}
    \text{RMSE} = \sqrt{\frac{\sum_{k=1}^{N_{h}} \lVert \hat{\mathbf{x}}_k - \tilde{\mathbf{x}}_k \rVert^2}{N_{h}}}
\end{gathered}
\end{equation}
where $N_h$ denotes the length of the prediction horizon, i.e., $N_h = 60$ in this case.

\begin{figure}[!tb]
\centering
\includegraphics[width=0.48\textwidth]{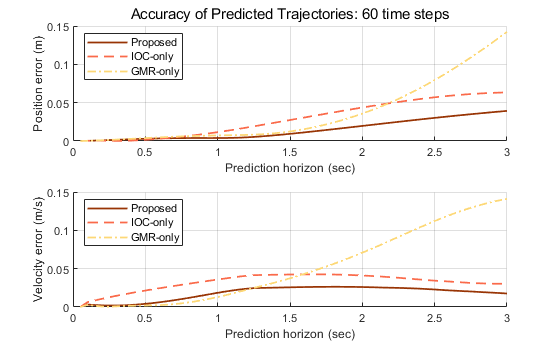}
\caption{Comparison of the trajectory prediction errors for a finite time-horizon. This is an illustrative result from a single testing data set in CS1.}
\label{fig_12}
\end{figure}

\begin{figure}[!tb]
\centering
\includegraphics[width=0.45\textwidth]{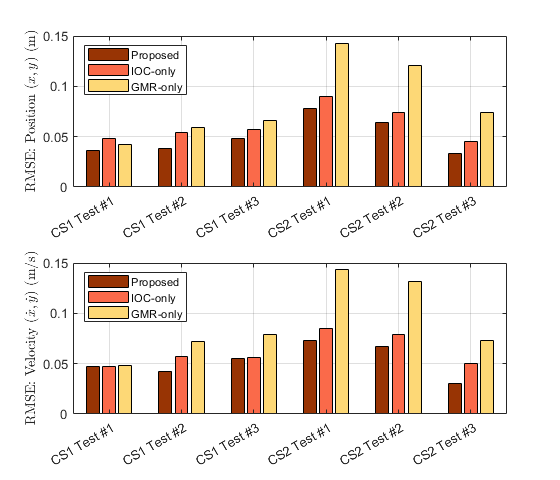}
\caption{(Top) The RMSE of the predicted position vector $[x_k \; y_k]^T$ and (Bottom) that of the predicted velocity vector $[\dot{x}_k \; \dot{y}_k]^T$ for 60 time steps between each method and the testing data.}
\label{fig_13}
\end{figure}

In Fig. \ref{fig_12}, the trajectory prediction errors using the three different methods are presented in a single testing data set from CS1. Figure \ref{fig_13} demonstrates the trajectory prediction accuracy of all testing data sets (three testing data sets from CS1 and three testing data sets from CS2). The RMSE of the predicted position vector $[x_k \; y_k]^T$ and that of the velocity vector $[\dot{x}_k \; \dot{y}_k]^T$ are presented separately. The proposed method provides the most accurate results compared with the other two methods in terms of the RMSE.
For the position error, the proposed method has $19.1\%$ and $40.8\%$ lower errors on average than the IOC-only and GMR-only, respectively. For the velocity error, the proposed method has $15.6\%$ and $42.5\%$ lower errors on average than the IOC-only and GMR-only, respectively.
The identified variability further improves the trajectory prediction accuracy over the IOC-only method. It is shown that the GMR-only method is data-inefficient; a possible explanation is that the GMR-only method cannot account for the structured task-objective-based behavior explicitly, which reduces the model accuracy when the amount of the training data is limited. On the other hand, the proposed method can predict human behavior more accurately, even with a small amount of data, by encoding the variability. The proposed method utilizes the IOC method to identify the task-objective-based behavior which dominates the modeled human behavior.

\subsection{Multiple-Subject Case Study}
We recruited 10 additional human-subjects to conduct a multiple-subject case study. All human-subjects were not exposed to the quadrotor simulation environment before the experiment. The main purpose is to show that the proposed method can account for different personal characteristics by providing customized models for each human-subject. Note that the only requirement for the human-subjects was to land the quadrotor consistently and safely without crashing. The consistency was requested to identify their personal behavioral patterns. The safety was required to meet Assumption \ref{assumption_K}. No specific control strategy was demanded.
\subsubsection{Hypothesis and Procedure}
A hypothesis to be tested is given as follows.
\begin{itemize}
    \item Hypothesis: The proposed method can predict future human behavior in the quadrotor landing scenario with higher accuracy compared to two baseline methods, the IOC-only and the GMR-only.
\end{itemize}

We provided basic information to all subjects regarding the experiment using the same material for about five minutes. The experiment procedure is composed of two phases: first, each subject is allowed to practice the quadrotor landing scenario with 10 minutes time limit. Their data is not recorded in this phase. A five-minute break follows. Second, each subject performs the landing mission 11 times. Their data is recorded in this phase. Among the recorded data, one trajectory is randomly chosen as testing data. The remaining 10 trajectories are used as training data. A prediction time-horizon is set to five seconds for the multiple-subject case study because the average time to land (about $15$ seconds) is slightly larger than the single-subject case study (about $12$ seconds).

\begin{figure}[!tb]
\centering
\includegraphics[width=0.45\textwidth]{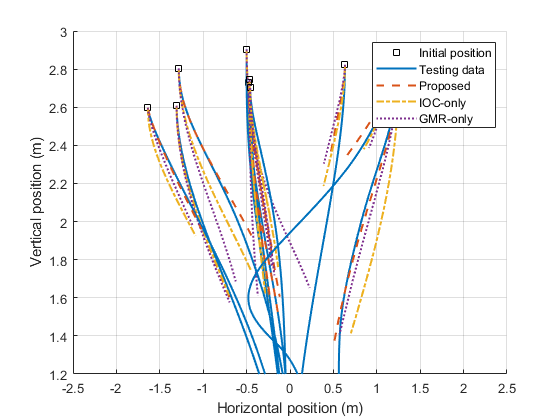}
\caption{The testing data in the multiple-subject case study and the predicted trajectories for 5 seconds from the initial conditions using each modeling method.}
\label{fig_mutiple}
\end{figure}

\begin{figure}[!tb]
\centering
\includegraphics[width=0.45\textwidth]{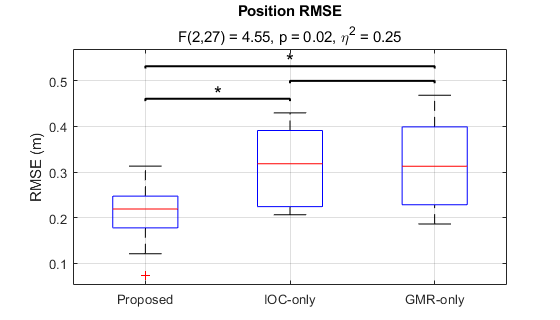}
\caption{The RMSE of position for 5 seconds between the predicted trajectory using each method and the testing data.}
\label{fig_box1}
\end{figure}

Figure \ref{fig_mutiple} shows 10 testing data (actual quadrotor trajectories in testing data of each human-subject) and the predicted quadrotor trajectories using the three modeling methods. Note that the quadrotor trajectories are predicted using 10 different human models (one model for each human-subject). In Fig. \ref{fig_box1}, the trajectory position prediction accuracy for each modeling method is presented as a box plot. We use the analysis of variance (ANOVA) for statistical testing \cite{Kalpic2011}. The ANOVA test reveals that there are significant differences between groups (modeling methods) in the position prediction accuracy ($F(2,27) = 4.55, p = 0.02, \eta^2 = 0.25$). We also used the pairwise T-tests for multiple groups. A conventional symbol ($*$) is used in Fig. \ref{fig_box1} to represent a significant p-value: $*p < 0.05$. The result shows that the proposed method is significantly more accurate in position prediction compared to the IOC-only and the GMR-only.

\begin{figure}[!tb]
\centering
\includegraphics[width=0.45\textwidth]{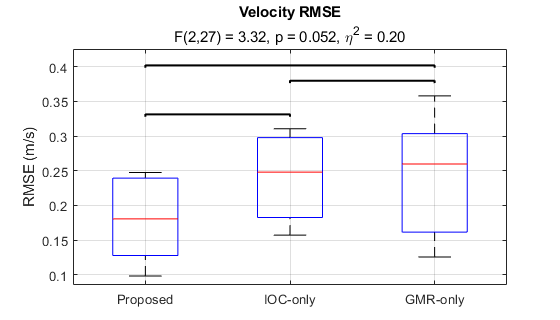}
\caption{The RMSE of velocity for 5 seconds between the predicted trajectory using each method and the testing data.}
\label{fig_box2}
\end{figure}

In Fig. \ref{fig_box2}, the RMSE of the predicted velocity is presented. The statistical testing results reveal that there are no significant differences between groups in the velocity prediction accuracy ($F(2,27) = 3.32, p = 0.052, \eta^2 = 0.20$). Nevertheless, the velocity prediction accuracy of the proposed method is improved by $26.3 \%$ and $27.5 \%$ compared to the IOC-only and GMR-only, respectively. We can explain this result: since the trained human behavior model utilized the system dynamics \eqref{eq:system_dynamics}, position, and velocity information, the prediction for the position is relatively accurate due to the imposed dynamic constraint. If one wants to predict velocity with higher accuracy, acceleration information can be measured and used. The system dynamics model needs to be extended accordingly. Then, the input to the GMM can incorporate acceleration information so that the velocity prediction accuracy can be enhanced.

\section{Conclusion} \label{Conlusion}
A human behavior modeling method that can account for not only the task objective but also the variability was proposed to describe and predict human behaviors. The proposed modeling method employed the inverse optimal control (IOC) approach to identify the task objective from the given human demonstrations. Then, the Gaussian mixture model (GMM) and Gaussian mixture regression (GMR) methods were used to estimate and parameterize the variability which is the uncertainty in human behavior and cannot be captured by the task objective. We demonstrated the efficacy of the proposed modeling method via human-subject experiments using a quadrotor landing scenario. The results showed that the proposed method can provide an explainable task objective function for the given human demonstrations and also infer the probabilistic distribution of the variability.
The prediction accuracy for human behavior was improved compared to the IOC-only method and the GMR-only method. The identified variability parameter can also provide a confidence level of the variability in terms of the covariance.



%

\appendix \label{Appendix}
The GMR-only method is used for the comparison study in Section \ref{Experiment}. This method encodes the human behavior $\mathbf{u}_k$ directly, instead of extracting the variability. Similar to  \eqref{eq:GMM_xi_mu}-\eqref{eq:GMM_Sigma}, the input and output of the conditional probability $p(\mathbf{u}_k \vert \mathbf{x}_k)$ are modeled as:
\begin{equation} \label{eq:GMM_only_xi_mu}
    \boldsymbol{\xi}'_k \triangleq \begin{bmatrix}
    \boldsymbol{\xi}_k^{'\mathcal{I}} \\ \boldsymbol{\xi}_k^{'\mathcal{O}}
    \end{bmatrix}
    = \begin{bmatrix}
    \mathbf{x}_k \\ \mathbf{u}_k
    \end{bmatrix}, \quad
    \bar{\boldsymbol{\mu}}^{'i} = \begin{bmatrix}
    \bar{\boldsymbol{\mu}}^{'i,\mathcal{I}} \\ \bar{\boldsymbol{\mu}}^{'i,\mathcal{O}}
    \end{bmatrix}
\end{equation}

\begin{equation} \label{eq:GMM_only_Sigma}
    \bar{\boldsymbol{\Sigma}}^{'i} = \begin{bmatrix}
    \bar{\boldsymbol{\Sigma}}^{'i,\mathcal{I}} & \bar{\boldsymbol{\Sigma}}^{'i,\mathcal{I} \mathcal{O}} \\
    \bar{\boldsymbol{\Sigma}}^{'i,\mathcal{O} \mathcal{I}} & \bar{\boldsymbol{\Sigma}}^{'i,\mathcal{O}}
    \end{bmatrix}
\end{equation}
Then, a set of GMM parameters $\{ \bar{h}^{'i}, \bar{\boldsymbol{\mu}}^{'i}, \bar{\boldsymbol{\Sigma}}^{'i} \}_{i=1}^G$ can be estimated using the EM algorithm. The estimated GMM parameters are exploited by the GMR method to compute the conditional probability $p(\mathbf{u}_k \vert \mathbf{x}_k)$ using  \eqref{eq:GMR_probability}-\eqref{eq:GMR_single_mu_sigma}, by replacing $\{ \bar{h}^{i}, \bar{\boldsymbol{\mu}}^{i}, \bar{\boldsymbol{\Sigma}}^{i} \}_{i=1}^G$ to $\{ \bar{h}^{'i}, \bar{\boldsymbol{\mu}}^{'i}, \bar{\boldsymbol{\Sigma}}^{'i} \}_{i=1}^G$. The task-parameterized GMM is learned as well, using \eqref{eq:state_frame}-\eqref{eq:TP_GMR}.



\section*{Acknowledgment}
The authors would like to acknowledge that this work is supported by
NSF CNS-1836952.

\ifCLASSOPTIONcaptionsoff
  \newpage
\fi



%

\bibliographystyle{IEEEtran}
\bibliography{IEEEabrv,mybib.bib}



%







\end{document}